\documentclass[10pt,twocolumn,letterpaper]{article}

\usepackage[pagenumbers]{cvpr} 

\usepackage[dvipsnames]{xcolor}

\usepackage{graphicx}
\usepackage{amsmath}
\usepackage{amssymb}
\usepackage{booktabs}
\usepackage{enumitem}
\usepackage{microtype}
\usepackage[dvipsnames]{xcolor}
\usepackage{bm, bbm}
\usepackage{algorithmic}
\usepackage[linesnumbered,boxed,ruled,commentsnumbered]{algorithm2e}
\usepackage{multirow}

\usepackage{makecell}

\usepackage[accsupp]{axessibility}

\usepackage[normalem]{ulem}
\useunder{\uline}{\ul}{}

\definecolor{cvprblue}{rgb}{0.21,0.49,0.74}
\usepackage[pagebackref,breaklinks,colorlinks,citecolor=cvprblue]{hyperref}

\def\paperID{6721} 
\def\confName{CVPR}
\def\confYear{2024}

\title{Enhancing 3D Object Detection with 2D Detection-Guided Query Anchors}

\author{Haoxuanye Ji$^{2,*,\sharp}$ \quad\quad Pengpeng Liang$^{1,*,\dagger}$ \quad\quad Erkang Cheng$^{2,\ddagger}$\\
$^1$School of Computer and Artificial Intelligence, Zhengzhou University, $^2$Nullmax \\
{jihaoxuanye@163.com, \{liangpcs, twokang.cheng\}@gmail.com}
\thanks{$^*$Equal contribution. $^\sharp$Work done during an internship at Nullmax.}
\thanks{$^\dagger$Project lead. $^\ddagger$Corresponding author.}
}

\begin{document}

\makeatletter
\def\thanks#1{\protected@xdef\@thanks{\@thanks
        \protect\footnotetext{#1}}}
\makeatother

\maketitle

\begin{abstract}
Multi-camera-based 3D object detection has made notable progress in the past several years. However, we observe that there are cases (e.g. faraway regions) in which popular 2D object detectors are more reliable than state-of-the-art 3D detectors. In this paper, to improve the performance of query-based 3D object detectors, we present a novel query generating approach termed QAF2D, which infers 3D \textbf{q}uery \textbf{a}nchors \textbf{f}rom \textbf{2D} detection results. A 2D bounding box of an object in an image is lifted to a set of 3D anchors by associating each sampled point within the box with depth, yaw angle, and size candidates. Then, the validity of each 3D anchor is verified by comparing its projection in the image with its corresponding 2D box, and only valid anchors are kept and used to construct queries. The class information of the 2D bounding box associated with each query is also utilized to match the predicted boxes with ground truth for the set-based loss. The image feature extraction backbone is shared between the 3D detector and 2D detector by adding a small number of prompt parameters. We integrate QAF2D into three popular query-based 3D object detectors and carry out comprehensive evaluations on the nuScenes dataset. The largest improvement that QAF2D can bring about on the nuScenes validation subset is $2.3\%$ NDS and $2.7\%$ mAP. Code is available at \url{https://github.com/nullmax-vision/QAF2D}.
\end{abstract}


\vspace{-8mm}

\section{Introduction}
\label{sec:intriduction}

3D object detection with multi-view images captured by surrounding cameras plays an important role in autonomous driving systems, and camera-based approaches have the benefit of low deployment cost in comparison to LIDAR-based approaches~\cite{eccv2022swformer, cvpr2023voxelnext, cvpr2022ess}. Though notable progress has been made in the past several years~\cite{eccv2020lss,eccv2022petr,iccv2023streampetr, aaai2023baseline,eccv2022bev,aaai2022polar},  multi-camera-based 3D object detection is still a challenging task due to the lack of depth information and the small object size in faraway regions.

\begin{figure}[t]
  \centering
   \includegraphics[width=1.\linewidth]{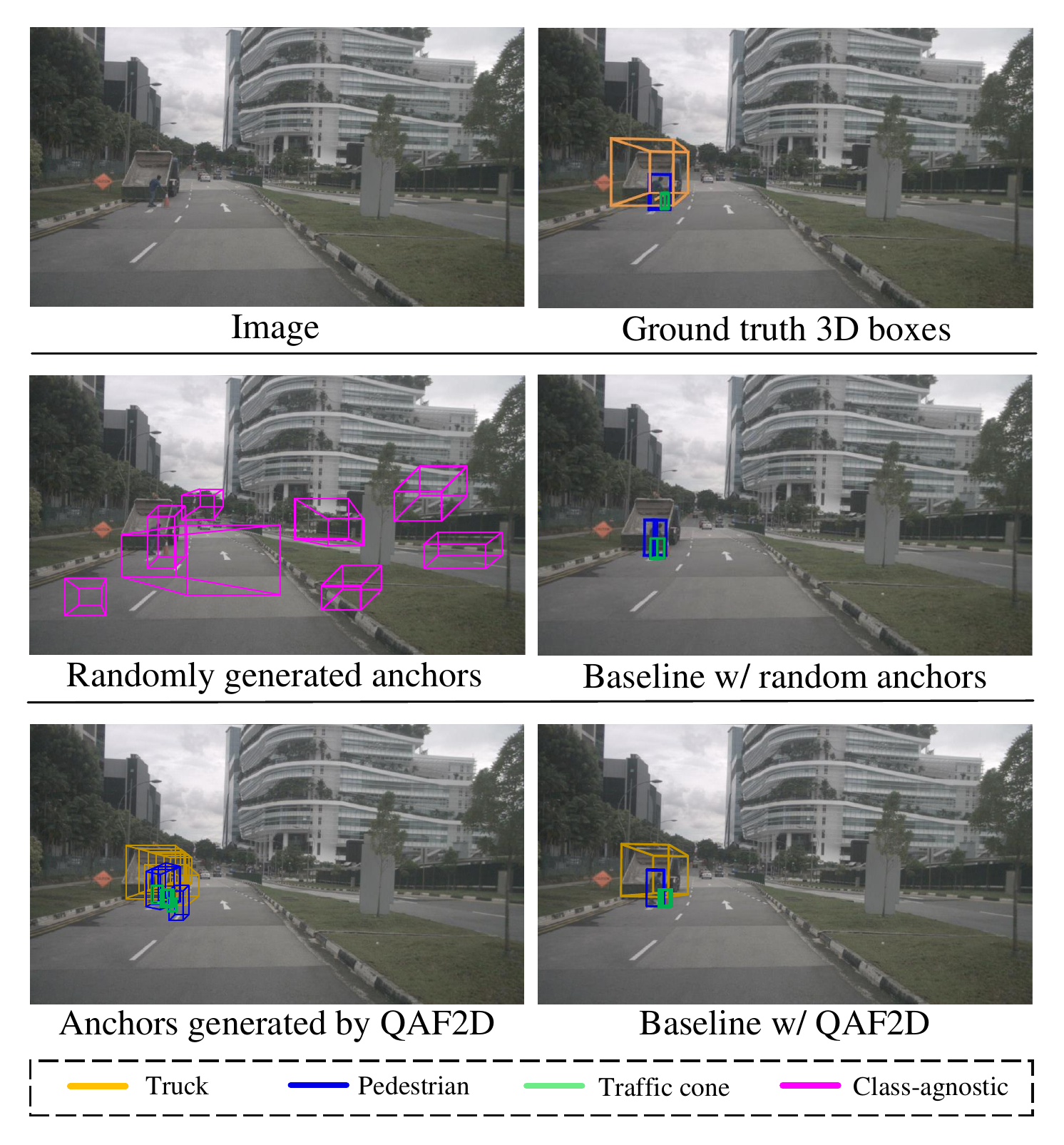}
   \caption{Comparison between randomly generated anchors and anchors generated by our QAF2D and comparison between their corresponding detection results. We use StreamPETR~\cite{iccv2023streampetr} as the baseline. Note that for illustration purpose, we just draw part of the anchors to alleviate clutter. }  
   \label{fig:motivation}
   \vspace{-5mm}
\end{figure}



Inspired by the promising performance of query-based 2D object detectors~\cite{eccv2020detr, iclr2022dab, iclr2020deformabledetr, cvpr2022dn, iccv2021conditiondetr}, query-based strategy~\cite{eccv2020detr} has been explored by several recent works for multi-camera-based 3D object detection~\cite{colr2021detr3d, iccv2023sparsebev, eccv2022petr, liu2022petrv2, iccv2023streampetr, eccv2022bev}. DETR3D~\cite{colr2021detr3d} projects a set of sparse 3D object queries to the 2D images for image feature aggregation. PETR~\cite{eccv2022petr, liu2022petrv2} constructs queries based on 3D points to interact with 3D position-aware image features. SparseBEV~\cite{iccv2023sparsebev} initializes a set of sparse queries based on pillars in bird's-eye-view (BEV) space which is used to sample  multi-view image features of several frames. StreamPETR~\cite{iccv2023streampetr} is built upon~\cite{eccv2022petr} with query propagation for temporal information modeling. BEVFormer~\cite{eccv2022bev} adopts a 3D detection head based on Deformable DETR~\cite{iclr2020deformabledetr} after constructing the BEV feature maps. Despite these approaches bringing about meaningful performance improvement, we observe there are some cases for which popular 2D detectors can handle successfully but 3D detectors fail. 

There are a few approaches~\cite{jiang2023far3d,wang2023mv2d,aaai2023baseline} trying to boost the 3D detection performance with the aid of 2D detectors. To generate 3D proposals, Far3D~\cite{jiang2023far3d} lifts the 2D detection bounding boxes to 3D with the depth estimated by a separate network. MV2D~\cite{wang2023mv2d} infers  3D reference points for object query generation by first transforming each 2D bounding box to a 2.5D point with RoI image features of the 2D box. A problem of~\cite{jiang2023far3d,wang2023mv2d} is that their approaches to lift 2D results  involve inferring depth information from images, which itself is a challenging task. SimMOD~\cite{aaai2023baseline} uses 2D bounding box detection as an auxiliary task during training to improve the perception of fine-grained structures, but it cannot use 2D detection results to directly provide guidance for 3D object detectors.

Considering the above problems, in this paper, we propose an approach named QAF2D to generate 3D query anchors from 2D detection bounding boxes to improve the performance of query-based 3D object detectors. More specifically, to lift the 2D bounding box of an object in an image to a set of 3D anchors, as the projection of the object's 3D center is within the 2D box, we first uniformly sample a set of projected centers inside the 2D box, and for each center, we associate it with depth, 3D size, and yaw angle candidates to generate 3D anchors. The set of 3D sizes chosen for each 2D detected box depends on the class of the box.  Then, each 3D anchor in the initial set is projected back to the image, and the IoU between the projected box and the corresponding 2D box is calculated, and only anchors having IoUs larger than a threshold are used to construct 3D queries. To make use of the class information of the 2D bounding boxes, for the calculation of the set-to-set loss based on DETR~\cite{eccv2020detr}, we associate each query with the class of its corresponding 2D box, and each predicted 3D bounding box can only be matched with ground truth boxes that have the same class. Fig.~\ref{fig:motivation} illustrates the advantage of the proposed anchor generation method.  

To reduce the computation cost while keeping the performance of the 3D detector as intact as possible, the image feature extraction backbone is shared by the 3D detector and 2D detector via prompt tuning~\cite{eccv2022vpt,arxiv2022prompt}, and the network is trained in two stages. In the first stage, the 3D detector is trained, and the 2D detection results are obtained by projecting the 3D ground truth to the multi-view images. In the second stage, only the prompt parameters and 2D detection head are trained with all other parameters frozen. We integrate the proposed QAF2D into three query-based 3D detectors (StreamPETR~\cite{iccv2023streampetr}, SparseBEV~\cite{iccv2023sparsebev}, and BEVFormer~\cite{eccv2022bev}) and carry out comprehensive experiments on the nuScenes dataset~\cite{cvpr2020nuscene}. The performance of all three query-based 3D detectors can be improved,  an average improvement of $1.18\%$ NDS and $1.74\%$ mAP is achieved on the nuScenes validation subset, and the largest improvement is $2.3\%$ NDS and $2.7\%$ mAP. 

The contributions of our paper are summarized as follows:
\begin{itemize}
    \item We propose to generate 3D query anchors from 2D bounding boxes so that the results of the more reliable 2D detector can be directly used to improve the 3D detection performance. 
    \item We share the image feature extraction backbone between the 3D and 2D detectors by visual prompts for efficiency and successfully train the network in two stages. 
    \item Consistent performance improvement is achieved on the nuScenes dataset when the proposed QAF2D is integrated into three query-based 3D object detectors, and it shows the effectiveness and generalization ability of our proposed approach.
\end{itemize}

\section{Related Work}
\label{sec:relatedwotk}
\subsection{Camera-based 3D Object Detection}
Camera-based 3D detectors aim to predict the 3D bounding boxes of objects in camera images. Some approaches~\cite{iccvw2021fcos3d, iccv2023monodetr, cvpr2021dle, cvpr2020monopair} focus on the monocular setting. FCOS3D~\cite{iccvw2021fcos3d} utilizes a fully convolutional single-stage network to regress 3D object information directly without using any 2D-3D correspondence priors.~\cite{cvpr2021dle} identifies the localization error as a key factor that constrains the detection performances and proposes strategies to alleviate it. MonoPair~\cite{cvpr2020monopair} leverages spatial constraints between paired objects to deal with occlusion. MonoDETR~\cite{iccv2023monodetr} enhances the vanilla Transformer with the contextual depth cues to guide the 3D detection process.

Recently, 3D object detection under the surrounding multiple-camera setting has attracted a lot of research efforts. One line of approaches~\cite{eccv2020lss, aaai2023bevdepth, aaai2023bevstreo,eccv2022bev, aaai2022polar, iccv2023dfa3d} transforms the image features to the BEV space with the help of depth estimation before applying a detection head. LSS~\cite{eccv2020lss} utilizes an estimated depth distribution of each pixel to lift the features of each image individually into a frustum and converts the frustums of all images into a BEV grid. BEVDepth~\cite{aaai2023bevdepth} uses the intrinsic camera parameters as one of the inputs of the depth estimation module with supervision from point cloud to predict depth of images for BEV feature construction. BEVStereo~\cite{aaai2023bevstreo} designs an effective depth estimation method based on temporal stereo to build BEV features. BEVFormer~\cite{eccv2022bev} proposes to aggregate features from both the spatial and temporal spaces to the current BEV space with learnable queries.  PolarFormer~\cite{aaai2022polar} builds the BEV features in the polar coordinate system to consider  the wedge shape of the physical world under  the ego car’s perspective. DFA3D~\cite{iccv2023dfa3d} proposes to use 3D deformable attention to aggregate the lifted features in 3D space so that the depth ambiguity problem can be mitigated. 

Another line of approaches~\cite{colr2021detr3d, liu2022petrv2, cvpr2023cape, iccv2023streampetr, wang2023mv2d} directly samples image features with queries and uses a decoder network to detect objects. DETR3D~\cite{colr2021detr3d} proposes to use a set of sparse object queries to implicitly transform features from 2D to 3D without estimating dense 3D scene geometry. PETR~\cite{eccv2022petr} generates  3D position-aware image features and then uses a set of queries to interact with the features and predict 3D bounding boxes. PETRv2~\cite{liu2022petrv2} constructs the 3D position embedding in a data-dependent way, and temporal information is exploited by transforming the coordinates in the previous frame to the current coordinate system. StreamPETR~\cite{iccv2023streampetr} makes use of temporal information by propagating selected queries from a memory queue to the current frame, and a motion-aware layer normalization is designed as well. SparseBEV~\cite{iccv2023sparsebev} lets the queries interact with image features in a sparse manner with promising performance by designing scale-adaptive self-attention and  adaptive spatio-temporal sampling modules.   MV2D~\cite{wang2023mv2d} proposes to learn queries based on 2D detection results, which later interact with RoI image features.  CAPE~\cite{cvpr2023cape} adopts a local camera-view coordinate system instead of a global one to form 3D position embeddings so that variances caused by changes of camera extrinsic parameters can be eliminated.


\subsection{Query-based 2D Object Detection}
DETR~\cite{eccv2020detr} is the first Transformer-based~\cite{iclr2021vit} object detection approach, and it uses a set of object queries to interact with images features and constructs loss via  bipartite matching. Many subsequent works~\cite{iclr2020deformabledetr, iclr2022dab, iccv2021conditiondetr, aaai2022anchor, cvpr2022dn, iccv2023stablematching} have proposed to improve DETR. Deformable-DETR~\cite{iclr2020deformabledetr} proposes deformable attention that only attends to a small number of key sampling points of a reference.  To accelerate the convergence, Conditional-DETR~\cite{iccv2021conditiondetr} disentangles the content and spatial queries and predicts conditional spatial queries from the decoder embedding. Anchor-DETR~\cite{aaai2022anchor} uses anchor points to build queries so that each query can focus on a specific region. DAB-DETR~\cite{iclr2022dab} proposes to construct queries with dynamic learnable anchor boxes which are updated layer-by-layer. DN-DETR~\cite{cvpr2022dn} reduces the instability of bipartite graph matching by reconstructing ground truth boxes from noisy queries.~\cite{iccv2023stablematching} proposes to only use positional metrics to stabilize the matching process of the DETR loss.

\subsection{Visual Prompt Tuning} 
Prompting is initially proposed to modify the input text string so that a pre-trained large language model can be adapted to new tasks with few or no labeled data~\cite{liu2023promptpre}. CLIP~\cite{icml2021visnlpprompt} uses prompts to transfer visual models trained with natural language supervision to downstream tasks under the zero-shot setting. Recently, some works~\cite{eccv2022vpt, iclr2023lpt, arxiv2022prompt, iccv2023e2vpt} propose to adapt a pre-trained visual model to different domains by adding a small amount of prompt parameters instead of fine-tuning the entire model. VPT~\cite{eccv2022vpt} utilizes a small amount of trainable parameters to adapt large-scale Transformer models to downstream tasks instead of full fine-tuning. LPT~\cite{iclr2023lpt} introduces a shared prompt and group-specific prompts into a frozen pre-trained model  to adapt to long-tailed data.~\cite{arxiv2022prompt} transforms the input image with prompts so that a frozen pre-trained model can perform new tasks.  E$^2$VPT~\cite{iccv2023e2vpt} proposes to use learnable key-value prompts and visual prompts with  a prompt pruning procedure for effective and efficient fine-tuning. 
\section{Method}
\label{sec:method}

\subsection{Overall Architecture}
\begin{figure*}[t]
  \centering
   \includegraphics[width=1.0\linewidth]{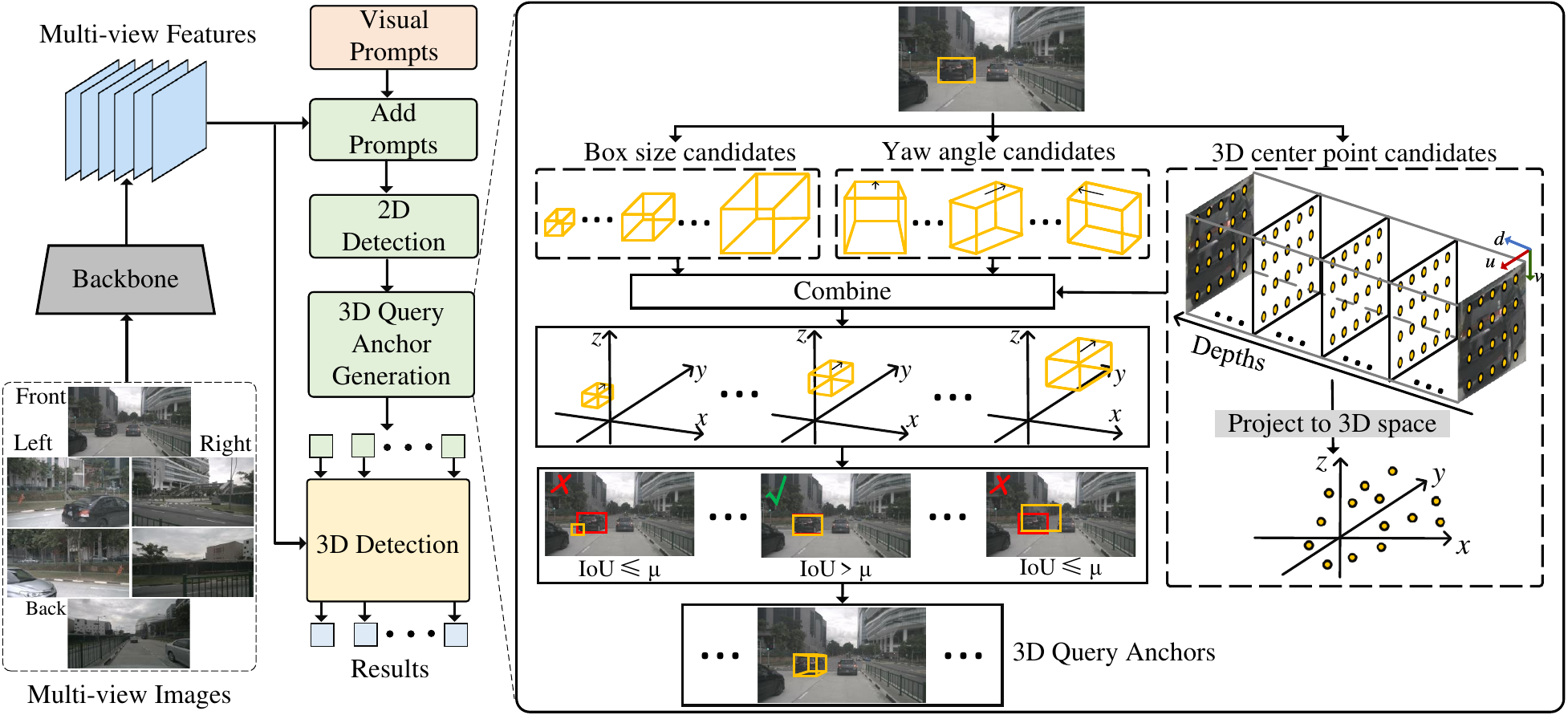}
   \caption{
   Overview of the 3D detection pipeline with our proposed 3D query anchor generation approach. The image backbone network extracts features of the input multi-view images, and the features are shared between the 3D detector and 2D detector with visual prompts. 2D detection results are used to generate 3D query anchors. Our 3D anchor generation method first generates box size candidates, yaw angle candidates, and 3D center point candidates, and then combines them to construct an initial set of anchors, which is refined with IoU check to form the final set of 3D query anchors. The entire network is optimized in two stages.}
   \label{fig:framework}
   \vspace{-3mm}
\end{figure*}

As shown in Fig.~\ref{fig:framework},  at timestamp $t$, the captured multi-camera images $I_t=\{I_c^t\}_{c=1}^{N_{\mathrm{cam}}}$ are input into an feature extraction backbone (e.g. ResNet~\cite{cvpr2016resnet} or VovNet~\cite{cvpr2019v2net}) to extract image features $F_t=\{F_c^t\}_{c=1}^{N_{\mathrm{cam}}}$, where $N_{\mathrm{cam}}$ is the number of cameras.  We first feed $F_t$ to the 2D detection branch to obtain 2D bounding boxes. The 3D anchor generator generates a set of 3D anchors for each 2D bounding box with its corresponding camera's intrinsic and extrinsic parameters and its class information. Then, the query-based 3D detector takes $F_t$ and the generated 3D anchors as input to predict 3D bounding boxes. We train the network in two stages. 


\subsection{3D Query Anchor Generation}
\label{sec:3d-query-anchors}

Given the 2D detection bounding boxes $B=\{(\mathbf{b}_i, g_i)\}_{i=1}^{N}$ of image $I$, where $\mathbf{b}_i=(x_i,y_i,w_i,h_i)$ represents a bounding box with its center coordinate, width, and height, and $g_i$ is its class information, we associate each sampled point in $\mathbf{b}_i$ with 3D size, depth and yaw angle candidates, and generate a set of 3D query anchors for $\mathbf{b}_i$ following the steps described below. The process is shown on the right side of Fig.~\ref{fig:framework}.  

\textbf{3D anchor center candidates}. Since the projection of the center of an object to an image plane is always within the minimum unrotated rectangle that contains the entire object, given a 2D bounding box   $\mathbf{b}_i=(x_i,y_i,w_i,h_i)$, we first sample a set of 2D object centers in the following way:
\begin{equation}
\begin{aligned}
\label{eq:c2d}
C_{\mathrm{2D}} &= \{(x_s, y_s)|x_s=x_{\mathrm{min}}+s_x\times i_x, \\
&y_s=y_{\mathrm{min}}+s_y\times i_y; \\ 
&i_x, i_y\in \mathbb{Z}_{\geq 0}, x_s\leq x_{\mathrm{max}}, y_s\leq y_{\mathrm{max}}\},
\end{aligned}
\end{equation}
where $s_x$ and $s_y$ are step sizes, $x_{\mathrm{min}}=\lfloor x_i-w_i/2\rfloor$, $x_{\mathrm{max}}=\lfloor x_i+w_i/2\rfloor$, $y_{\mathrm{min}}=\lfloor y_i-h_i/2\rfloor$, and $y_{\mathrm{max}}=\lfloor y_i+h_i/2\rfloor$. Then, we define a set of depth candidates $D=\{d_0,...,d_{N_D-1}\}$ of size $N_D$ and associate each point in $C_{\mathrm{2D}}$ with the depth candidates to generate $C'_{\mathrm{2D}} = \{(x_s, y_s, d_s)|(x_s,y_s)\in C_{\mathrm{2D}},d_s\in D\}$. Next, we transform each point in $C'_{\mathrm{2D}}$ to the 3D coordinate system and obtain a 3D object center set $C_{\mathrm{3D}}=\{(x_e,y_e,z_e)\}$. The transformation is carried out with the intrinsic parameters $\mathbf{K}$ and extrinsic parameters $\mathbf{R}$ of the corresponding camera of $(x_s,y_s,z_s)$ using the formula below:
\begin{equation}
     \left[\begin{array}{c} x_e \\ y_e \\ z_e \end{array}\right]=\mathbf{R}^{-1}\mathbf{K}^{-1}\left[\begin{array}{c} x_s\cdot d_s \\ y_s\cdot d_s \\ d_s \end{array}\right]
\end{equation}

\textbf{3D anchor size candidates}. As the 3D object sizes of different classes can vary considerably  while the change of object size of the same class is usually small, we select a bounding box $\mathbf{b}_i$'s 3D size candidates based on its class $g_i$. For each class $g$, we decide its width range $(w_{\mathrm{g}}^{\mathrm{min}},w_{\mathrm{g}}^{\mathrm{max}})$, height range $(h_{\mathrm{g}}^{\mathrm{min}},h_{\mathrm{g}}^{\mathrm{max}})$, and length range $(l_{\mathrm{g}}^{\mathrm{min}},l_{\mathrm{g}}^{\mathrm{max}})$ by traversing all objects of class $g$ in the training data and choosing the maximum and minimum values. Based on the above ranges, we construct the sets of width, height, and length as follows:
\begin{equation}
\begin{aligned}
\label{eq:whl-sets}
W_g&=\{w | w=w_{\mathrm{g}}^{\mathrm{min}}+s_w\times i; i\in \mathbb{Z}_{\geq 0}, w\leq w_{\mathrm{g}}^{\mathrm{max}}\},\\
H_g&=\{h |h=h_{\mathrm{g}}^{\mathrm{min}}+s_h\times i; i\in \mathbb{Z}_{\geq 0}, h\leq h_{\mathrm{g}}^{\mathrm{max}}\},\\
L_g&=\{l |l=l_{\mathrm{g}}^{\mathrm{min}}+s_l\times i; i\in \mathbb{Z}_{\geq 0}, l\leq l_{\mathrm{g}}^{\mathrm{max}}\}
\end{aligned}
\end{equation}
where $s_w$, $s_h$, $s_l$ are the step sizes. Then, the set of 3D object size candidates $S_g$ for class $g$ are generated by combining the above three sets:
\begin{equation}
\label{eq:boxsize}
S_g=\{(w,h,l)|w\in W_g, h\in H_g, l\in L_g\}
\end{equation}

\textbf{Yaw angle candidates}. We construct the set of yaw angle candidates by uniformly sampling in $[0, 2\pi)$ with interval $\pi/N_{\theta}$, and the set $\Theta$ is defined below:
\begin{equation}
\begin{aligned}
\label{eq:yaw}
\Theta=\{\theta|\theta=\frac{n_{\theta}}{N_{\theta}}\pi, n_{\theta}=0,1,2,...,2N_{\theta}-1\}
\end{aligned}
\end{equation} 

\textbf{Generating 3D query anchors.} Given the 3D center candidate set $C_{\mathrm{3D}}$, size candidate set $S_{\mathrm{g}}$, and yaw angle candidate set $\Theta$ of a 2D bounding box $\mathbf{b}$ of class $g$, we generate an initial set of 3D anchors via the Cartesian product of the three candidate sets, i.e. $P_\mathrm{Init}=\{\mathbf{p}_i|\mathbf{p}_i=(x_i,y_i,z_i,w_i,h_i,l_i,\theta_i)\}=C_{\mathrm{3D}}\times S{\mathrm{g}}\times \Theta$. To remove the anchors that are not compatible with $\mathbf{b}$, we project each $\mathbf{p}_i$ to the image plane of the 2D bounding box  $\mathbf{b}$ and get $\mathbf{p}_i^{\mathrm{2D}}=(x_i^{\mathrm{2D}},y_i^{\mathrm{2D}},w_i^{\mathrm{2D}},h_i^{\mathrm{2D}})$. Then, the IoU between $\mathbf{p}_i^{\mathrm{2D}}$ and $\mathbf{b}$ is calculated, and only $\mathbf{p}_i$ with IoU larger than the threshold $\mu$ is retained. The final set of query anchors is $P=\{\mathbf{p}_i|\mathrm{IoU}(\mathbf{p}_i^{\mathrm{2D}},\mathbf{b}))>\mu$\}. We use $P$ as the decoder input of three selected query-based 3D object detectors to show its effectiveness. 

\subsection{Two-stage Optimization with Visual Prompts}
\label{subsec:optimize}
To enable the 2D detection branch (DAB-DETR~\cite{iclr2022dab} is used as the default 2D detector) to share the image feature extraction backbone with the 3D detection branch without compromising the performance of the 3D detector, we train the network in two stages with visual prompts~\cite{arxiv2022prompt} added to the 2D detection branch. 

\textbf{Stage 1: Training 3D detection branch.} For training, instead of using the output of the 2D detection branch, we project the ground truth 3D bounding boxes to the images to get 2D bounding boxes. Then, the proposed 3D query anchor generation method is used to generate 3D query anchors which are used as the input to the decoder of the query-based 3D object detector (based on StreamPETR~\cite{iccv2023streampetr}, SparseBEV~\cite{iccv2023sparsebev}, or BEVFormer~\cite{eccv2022bev}). To take advantage of the class information of 2D bounding boxes, when calculating the set-to-set loss~\cite{eccv2020detr} of the query-based 3D detector, the predicted 3D bounding box of a query is only matched with a ground truth box that has the same class as the query's corresponding 2D box. More specifically, we divide the set of predicted bounding boxes $\hat{B}$ and the set of ground truth bounding boxes $B$ into $\{\hat{B}_g\}_{g=1}^{G}$ and $\{B_g\}_{g=1}^{G}$, respectively, where $G$ is the number of classes. Given a set pair $(\hat{B}_g,B_g)$,  boxes in $\hat{B}_g$ are matched with boxes in $B_g$ by Hungarian algorithm following~\cite{eccv2020detr},  and the cost matrix is calculated based on the predicted probability of the ground truth class and the $L_1$ loss between the predicted and ground truth 3D bounding boxes. 

\textbf{Stage 2: Training 2D detection branch with visual prompts.} After training the image feature backbone with the 3D detection task in Stage 1, the backbone is frozen, and the 2D detection branch uses the same image features as the 3D detector. To adapt the image features to the 2D detection task, we add visual prompts designed in~\cite{arxiv2022prompt} to the feature maps. It is worth noting that ~\cite{arxiv2022prompt} adds prompts to the input images instead of feature maps as do we.  We adopt the padding prompt design of~\cite{arxiv2022prompt}. More specifically,  given a feature map of size $C\times H\times W$, where $C$, $H$, and $W$ are the number of channels, height, and width, respectively. Two prompt patches of size $C\times (\tau\times H)\times W$ are added to the top and bottom of the feature map, respectively. Another two prompt patches of size $C\times (H - 2\tau\times H)\times (\tau\times W)$ are added to the left and right sides of the feature map, respectively. The total number of prompt parameters varies with the change of $\tau$. For the training of the 2D detection branch, only the prompt parameters and the head of the 2D detector are updated. 

In the test phase, the 2D bounding boxes predicted by the 2D detection branch are used to generate the 3D query anchors for the query-based 3D detector. 

\subsection{Integrating into Query-based 3D Detectors}
\label{subsec:incorporate}

We integrate our approach into three selected query-based 3D object detectors (StreamPETR~\cite{iccv2023streampetr}, SparseBEV~\cite{iccv2023sparsebev}, BEVFormer~\cite{eccv2022bev}) by replacing the randomly initialized anchors (or learnable queries) with our proposed 3D query anchors inferred from 2D bounding boxes. 

\textbf{StreamPETR.} StreamPETR~\cite{iccv2023streampetr} is built upon the query-based 3D object detector PETR~\cite{eccv2022petr}. To make use of temporal information efficiently, it maintains a memory queue of historical object queries. The queries of the current frame consist of selected queries from the memory queue and newly added queries. The new queries depend on a set of learnable 3D anchor points initialized with uniform distribution between 0 and 1. We integrate the proposed 3D query anchors into StreamPETR by simply substituting a set of anchors $\{(x_i,y_i,z_i,w_i,l_i,h_i,\sin\theta_i,\cos\theta_i)\}$ inferred from 2D bounding boxes for  the original learnable 3D anchor points. 

\textbf{SparseBEV.} Besides adopting sparse queries, SparseBEV~\cite{iccv2023sparsebev} removes the dense global attention between queries and image features of~\cite{eccv2022petr}, and it proposes an adaptive spatio-temporal sampling method to aggregate image features. SparseBEV defines a set of learnable queries, and each query represents an object's translation, dimension, rotation, and velocity. To integrate our 3D query anchors, we replace the learnble queries with $\{(x_i,y_i,z_i,w_i,l_i,h_i,\sin\theta_i,\cos\theta_i)\}$ generated from 2D detection results. 

\textbf{BEVFormer.} In~\cite{eccv2022bev}, BEV features are constructed by a spatial cross-attention between predefined BEV queries and multi-camera image features, which  are used as input to a modified Deformable DETR~\cite{iclr2020deformabledetr} for 3D object detection. To apply the 3D query anchors, we design a 3D detection head based on DAB-Deformable-DETR~\cite{iclr2022dab}, and the learnable dynamic anchors in the form of $(x,y,h,w)$ are replaced with 3D query anchors in the form of $(x,y,z,w,l,h,\sin\theta,\cos\theta)$, and the decoder predicts 3D bounding boxes and velocity rather than 2D bounding boxes.
\section{Experiments}

\begin{table*}[t]
\centering
\setlength{\tabcolsep}{.31em}%
\begin{tabular}{l|c|c|cc|ccccc}
\hline
Methods                            & Backbone                       & Image Size & NDS           & mAP           & mATE & mASE & mAOE & mAVE & mAAE \\ \hline
StreamPETR \cite{iccv2023streampetr}                        &  & 320×800    & 57.1          & 48.2          & 61.0 & 25.6 & 37.5 & 26.3 & 19.4 \\
StreamPETR-8DQuery                       & V2-99 & 320×800    & 57.6          & 48.6          & 58.9 & 25.7 & 37.7 & 25.5 & 19.5 \\
StreamPETR-QAF2D (Ours)                  &                                & 320×800    & \textbf{58.6}          & \textbf{50.0}          & 56.1 & 26.1 & 36.9 & 25.1 & 19.6 \\
\hline
StreamPETR \cite{iccv2023streampetr}                        &  & 320×800    & 54.0 & 43.2 & 58.1 & 27.2 & 41.3 & 29.5 & 19.5 \\
StreamPETR-8DQuery                       & ResNet50 & 320×800    & 54.2 & 44.0 & 62.1 & 27.1 & 41.1 & 26.5 & 21.0 \\
StreamPETR-QAF2D (Ours)                  &                                & 320×800    & \textbf{54.6}  & \textbf{44.7} & 62.3 & 26.9 & 41.0 & 27.7 & 19.5 \\
\hline
StreamPETR$^{*\ddagger}$ \cite{iccv2023streampetr}                        &  & 320×800    & 55.0          & 45.0          & 61.3 & 26.7 & 41.3 & 26.5 & 19.6 \\
StreamPETR$^{*\ddagger}$-8DQuery                       & ResNet50 & 320×800    & 55.2          & 45.5          & 61.0 & 27.1 & 40.1 & 27.6 & 20.1 \\
StreamPETR$^{*\ddagger}$-QAF2D (Ours)                  &                       & 320×800    & \textbf{56.2}          & \textbf{46.5}          & 61.5 & 26.2 & 36.0 & 26.5 & 19.9 \\
\hline
SparseBEV \cite{iccv2023sparsebev}                         & \multirow{2}{*}{ResNet50}      & 256×704    & 55.8          & 44.8          & 58.1 & 27.1 & 37.3 & 24.7 & 19.0 \\
SparseBEV-QAF2D (Ours)                   &                         & 256×704    & \textbf{56.1} & \textbf{46.0} & 57.3 & 26.3 & 38.7 & 27.6 & 19.1 \\
\hline
BEVFormer-small \cite{eccv2022bev}  & \multirow{3}{*}{ResNet101-DCN}  & 736×1280 & 47.9 & 37.0          & 72.1 & 28.0 & 40.7 & 43.6 & 22.0 \\
BEVFormer-small-DAB3D           &  & 736×1280   & 49.2          & 39.0          & 71.7 & 27.5 & 41.6 & 42.2 & 19.7 \\
BEVFormer-small-QAF2D (Ours)             &                                & 736×1280   & \textbf{50.2}  & \textbf{39.7}  & 70.3 & 27.4 & 36.9 & 40.4 & 21.3 \\
\hline
\end{tabular}
\caption{Comparison of the base detectors and their QAF2D enhanced version on the nuScenes validation split. $*$ indicates the use of perspective-view pre-training. $\ddagger$ represents the use of 300 randomly initialized queries (irrelevant to QAF2D) and 128 propagation queries. Please refer to the corresponding text in Sec.~\ref{sec:effectiveness-qaf2d} for the meaning of ``8DQuery" and ``DAB3D". The best is in \textbf{bold}.}
\label{tab:val}
\vspace{-2mm}
\end{table*}

\begin{table*}[t]
\centering
\setlength{\tabcolsep}{.47em}%
\begin{tabular}{l|c|c|cc|ccccc}
\hline
Method                       & Backbone & Image Size & NDS  & mAP  & mATE & mASE & mAOE & mAVE & mAAE \\ \hline
DETR3D~\cite{colr2021detr3d}                       & V2-99    & 900×1600   & 47.9 & 41.2 & 64.1 & 25.5 & 39.4 & 84.5 & 13.3 \\
BEVFormer~\cite{eccv2022bev}                    & V2-99    & 900×1600   & 56.9 & 48.1 & 58.2 & 25.6 & 37.5 & 37.8 & 12.6 \\
PolarFormer~\cite{aaai2022polar}                  & V2-99    & 900×1600   & 57.2 & 49.3 & 55.6 & 25.6 & 36.4 & 43.9 & 12.7 \\
PETRv2~\cite{liu2022petrv2}                       & V2-99    & 640×1600   & 58.2 & 49.0 & 56.1 & 24.3 & 36.1 & 34.3 & 12.0 \\
CAPE~\cite{cvpr2023cape}                         & V2-99    & 640×1600   & 61.0 & 52.5 & 50.3 & 24.2 & 36.1 & 30.6 & 11.4 \\
MV2D~\cite{wang2023mv2d}                         & V2-99    & --         & 59.6 & 51.1 & 52.5 & 24.3 & 35.7 & 35.7 & 12.0 \\
SparseBEV~\cite{iccv2023sparsebev}                    & V2-99    & 640×1600   & 62.7 & 54.3 & 50.2 & 24.4 & 32.4 & 25.1 & 12.6 \\
SparseBEV~(dual-branch)~\cite{iccv2023sparsebev}       & V2-99    & 640×1600   & 63.6 & 55.6 & 48.5 & 24.4 & 33.2 & 24.6 & 11.7 \\ \hline
StreamPETR~\cite{iccv2023streampetr}                   & V2-99    & 640×1600   & 63.6 & 55.0 & 47.9 & 23.9 & 31.7 & 24.1 & 11.9 \\
StreamPETR-8DQuery                   & V2-99    & 640×1600   & 63.6  & 55.5  & 47.1  & 23.6  & 32.2  & 26.8  & 11.8  \\
StreamPETR-QAF2D~(Ours) & V2-99    & 640×1600   & \textbf{64.2} & \textbf{56.6} & 46.1 & 24.0 & 32.6 & 26.1 & 12.1 \\ \hline
\end{tabular}
\vspace{-1mm}
\caption{Comparison with the state-of-the-art approaches on the nuScene test split. The best is in \textbf{bold}.}
 \vspace{-5mm}
\label{tab:test}
\end{table*}

\subsection{Dataset and Metrics}
We conduct experiments on the nuScene dataset~\cite{cvpr2020nuscene}. It consists of 1000 multi-modal videos each of which is about 20s, and keyframes are annotated every 0.5s. The sensors include camera, LIDAR, and RADAR, and we use the images captured by the six surrounding cameras for our experiments. The videos are split into three subsets of 750, 150, and 150 for training, validation, and test, respectively. There are 1.4M annotated 3D bounding boxes of 10 classes in total. 

We use the official evaluation metrics of nuScenes. Along with mean average precision (mAP), the following true positive errors are reported: average translation error (ATE), average scale error (ASE), average orientation error (AOE), average velocity error (AVE), and average attribute error (AAE). In addition, a more comprehensive nuScenes detection score (NDS) is derived from the above metrics. 

\subsection{Implementation Details} 
\vspace{-1.5mm}
We implement our method with PyTorch~\cite{nips2017pytorch}. After integrating our QAF2D into a base detector, we use the base detector's data augmentation strategy and training setting (e.g. learning rate, batch size, number of epochs) for training. All models are trained with 8 NVIDIA GeForce RTX 3090 GPUs.  For the parameters regarding 3D anchors generation in Section~\ref{sec:3d-query-anchors}, $s_x$ and $s_y$ that control the sampling intervals of 2D object centers are set to 10. The depth candidates in $D$ are sampled  between 3 meters and 103 meters with an interval of 1.5 meters. The width range, height range and length range of each class are given in the appendix. We set the sampling intervals $s_w$, $s_h$ and $s_l$  to 5, and $N_\theta$ that controls the  sampling interval of yaw angle is set to 12. The IOU threshold $\mu$ for anchor validation  is 0.99, i.e. only 3D anchors, the projections of which have high overlap with their corresponding 2D boxes, are kept. The parameter $\tau$ related to the number of prompt parameters in Section~\ref{subsec:optimize} is set to 0.2 based on the ablation study.

\subsection{Effectiveness of QAF2D}
\label{sec:effectiveness-qaf2d}
\vspace{-1.5mm}
\quad \textbf{Validation on nuScenes val split.} To verify the effectiveness of the 3D query anchors generated by the proposed QAF2D, we compare StreamPETR~\cite{iccv2023streampetr}, SparseBEV~\cite{iccv2023sparsebev}, and BEVFormer~\cite{eccv2022bev} with their QAF2D-enhanced version on the nuScenes validation split. For fair comparison of StreamPETR, besides its original results using three dimensional random queries, we also report the results of StreamPETR with eight dimensional random queries (8DQuery), which have the same dimension as queries of QAF2D. For fair comparison of BEVFormer-small,  along with its original result based on modified Deformable DETR~\cite{iclr2020deformabledetr}, we report the result of BEVFormer-small with our modified  DAB-Deformable-DETR~\cite{iclr2022dab} (BEVFormer-small-DAB3D) as well. BEVFormer-small-DAB3D uses eight dimensional randomly initialized queries. For SparseBEV, as its queries are nine dimensional containing velocity information, we do not change queries to eight dimensional ones. The results in Table~\ref{tab:val} demonstrate that our QAF2D can bring about consistent improvement for all three base detectors. With regard to StreamPETR (V2-99 backbone~\cite{cvpr2019v2net}), after improving the performance of StreamPETR by $0.5\%$ NDS and $0.4\%$ mAP with eight dimensional queries, our proposed QAF2D can obtain additional improvement of $1.0\%$ NDS and $1.4\%$ mAP in comparison with StreamPETR-8DQuery, and the entire improvement is $1.5\%$ NDS and $1.8\%$ mAP. We also evaluate StreamPETR and its enhanced version with ResNet50 backbone~\cite{cvpr2016resnet}, and our QAF2D can gain an improvement of $0.6\%$ NDS and $1.5\%$ mAP. In addition, we incorporate QAF2D into StreamPETR$^{*\ddagger}$ (benefits from perspective-view pre-training and uses different number of queries~\cite{iccv2023streampetr}), and QAF2D can improve its performance by $1.2\%$ NDS and $1.5\%$ mAP.  For SparseBEV, our proposed QAF2D can improve its performance by $0.3\%$ NDS and $1.2\%$ mAP. As to BEVFormer, it can be first improved with our modified  DAB-Deformable-DETR by $1.3\%$ NDS and $2.0\%$ mAP, and the proposed OAF2D can gain another improvement of $1.0\%$ NDS and $0.7\%$ mAP compared with BEVFormer-small-DAB3D, and the total improvement is $2.3\%$ NDS and $2.7\%$ mAP.

\textbf{Comparison with state-of-the-art on nuScenes test split.}  We compare StreamPETR-QAF2D with the state-of-the-art approaches on the nuScenes test split. The results in Table~\ref{tab:test} show that while  StreamPETR-8DQuery has the same performance as StreamPETR in terms of NDS and mAP, our OAF2D still can improve StreamPETR by $0.6\%$ NDS and $1.6\%$ mAP, which further validates the effectiveness of our approach.  Meanwhile, QAF2D-enhanced StreamPETR achieves the best performance on the nuScenes test split. 

The performance improvement can be attributed to two aspects: (1) 3D query anchors inferred from 2D detection results can provide better initial 3D box positions and sizes than random anchors, and this can ease the optimization of the network and predict the results more accurately, (2) the state-of-the-art 2D detectors are more reliable than the 3D detectors, and some missed detections of the base 3D detectors can be recovered with the help of QAF2D. 


\begin{table}[t]
\centering
\resizebox{1.0\columnwidth}{!}{
\setlength{\tabcolsep}{0.1em}
\begin{tabular}{l|c|c|c|c|c||c}
\hline
          & Backbone & \makecell{2D \\ detection} & \makecell{3D anchor \\generation} & \makecell{3D \\ detection} & Total & Speed  \\ \hline
StreamPETR       & 45ms   & --        & --                   & 8ms    & 53ms 
 & 18.9 FPS\\
StreamPETR-QAF2D & 47ms   & 12ms    & 1ms              & 5ms    & 65ms & 15.4 FPS\\ \hline
\end{tabular}
}
\vspace{-2mm}
\caption{Component time consumption and speed comparison between StreamPETR  and StreamPETR-QAF2D with V2-99 backbone on an NVIDIA 3090 GPU.}
\label{tab:speed}
\vspace{-6mm}
\end{table}

\textbf{Component time consumption and speed.} We report the time consumption of each component and the speed of StreamPETR~\cite{iccv2023streampetr} and StreamPETR-QAF2D in Table~\ref{tab:speed}. Note that the two approaches use the same backbone, the slight difference in time between two separate runs should be inevitable for the hardware. Though StreamPETR-QAF2D is somewhat slower than StreamPETR (15.4 FPS vs 18.9 FPS), we think that the overall efficiency of  StreamPETR-QAF2D is acceptable. Meanwhile, the increase of the complexity mainly comes from the 2D detection head of DAB-DETR~\cite{iclr2022dab} and the 3D query anchor generation component is very fast (1ms). As QAF2D is not sensitive to the choice of the 2D detector (please refer to the ablation study in Section~\ref{subsec:abla}), the efficiency of QAF2D can be improved by using lighter 2D detectors.

\begin{table}[t]
\centering
\setlength{\tabcolsep}{0.125em}%
\begin{tabular}{l |cc }
\hline
Method              & NDS  & mAP  \\ \hline
BEVFormer-small-DAB3D    & 49.2 & 39.0 \\ \hline
BEVFormer-small-QAF2D (Faster-RCNN~\cite{pami2017faster}) & 50.0 & 39.5 \\
BEVFormer-small-QAF2D (DAB-DETR~\cite{iclr2022dab}) & \textbf{50.2}  & \textbf{39.7}  \\ \hline
\end{tabular}
\vspace{-2mm}
\caption{Comparison of QAF2D with different 2D detectors.}
\label{tab:queryanchors}
\vspace{-2mm}
\end{table}

\begin{table}[t]
\centering
\setlength{\tabcolsep}{0.3em}%
\begin{tabular}{l |cc }
\hline
Method              & NDS  & mAP  \\ \hline
BEVFormer-small-QAF2D & 50.2  & 39.7  \\ 
BEVFormer-small-QAF2D (w/ RA) & \textbf{50.3}  & \textbf{40.1}  \\ \hline
\end{tabular}
\vspace{-2mm}
\caption{Impact of additional random anchors (RA).}
\label{tab:learnable}
\vspace{-2mm}
\end{table}

\begin{table}[t]
\centering
\setlength{\tabcolsep}{0.2em}%
\begin{tabular}{c|cc|c}
\hline
 & NDS  & mAP  & \multicolumn{1}{l}{\# of prompt params.} \\ \hline
No sharing          & 50.3 & 40.2 & --  \\ \hline
Sharing w/o prompt      & 49.7  & 39.2  & --  \\ \hline
$\tau=0.1$    & 50.0 & 39.5 & 0.08M  \\
$\tau=0.2$    & \textbf{50.2} & \textbf{39.7} & 0.15M  \\
$\tau=0.3$    & 50.1  & \textbf{39.7}  & 0.20M  \\
$\tau=0.4$    & 49.8  & 39.4  & 0.23M   \\
$\tau=0.5$    & \textbf{50.2}  & 39.6  & 0.26M  \\ \hline
\end{tabular}
\vspace{-2mm}
\caption{Effect of visual prompts in feature sharing  and comparison of different $\tau$s.}
\label{tab:prompts}
\vspace{-6mm}
\end{table}

\subsection{Ablation Study}
\label{subsec:abla}
We carry out ablation study on the nuScenes validation split and use BEVFormer~\cite{eccv2022bev} as the base detector. 

\textbf{Generalization to different 2D detectors.} To study the generalization ability of QAF2D to different 2D detectors, besides the default 2D detector DAB-DETR~\cite{iclr2022dab} , we also combine QAF2D with the popular Faster-RCNN~\cite{pami2017faster} and applies it to BEVFormer~\cite{eccv2022bev}. Based on the results in Table~\ref{tab:queryanchors}, we can see that the performances of DAB-DETR and Faster-RCNN are very close to each other, which can demonstrate that the proposed QAF2D is not sensitive to the selection of the 2D detector. As the result of DAB-DETR is slightly better, we use DAB-DETR as the default 2D detector to provide 2D detection boxes. The insensitivity should be because that QAF2D does not need very precise 2D boxes, as long as the 2D detector does not miss objects, QAF2D can use rough 2D boxes to generate meaningful 3D query anchors that are better than random ones.



\begin{figure*}[t]
  \centering
   \includegraphics[width=1.0\linewidth,height=0.53\linewidth]{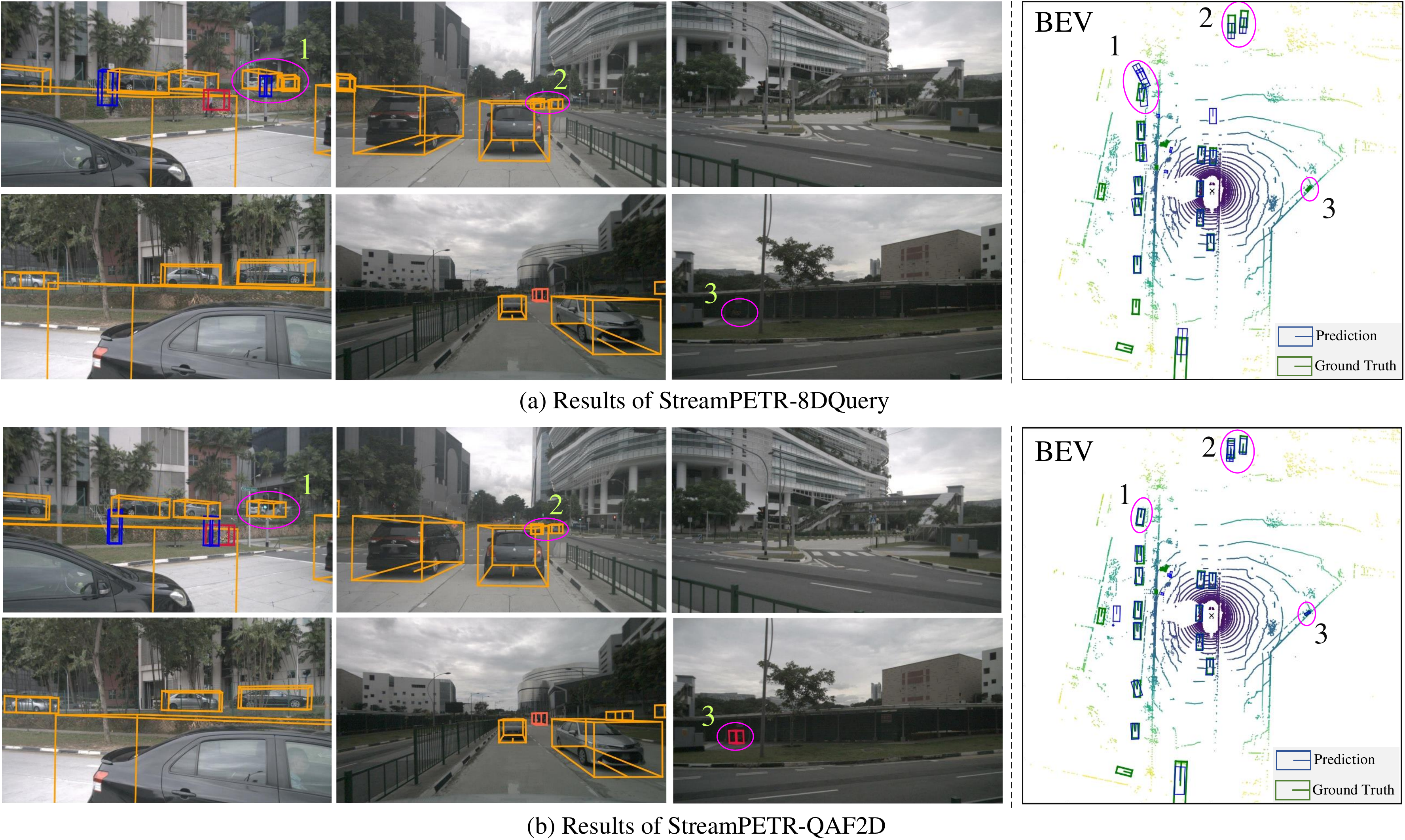}
   \vspace{-7.mm}
   \caption{Visualization results of StreamPETR-8DQuery and StreamPETR-QAF2D. The results in multi-camera images are shown on the left, and the corresponding results in bird's-eye-view are shown on the right. Three typical cases where StreamPETR-8DQuery fails but its QAF2D-enhanced version succeeds are in purple ellipses with numbers.}
   \label{fig:visualization}
   \vspace{-5mm}
\end{figure*}

\textbf{Impact of additional random 3D query anchors.} We also combine an additional set of randomly  initialized 3D query anchors with anchors generated by QAF2D to explore its impact. The number of random 3D query anchors is 900, which is the number of queries used by BEVFormer-small.  The results are shown in Table~\ref{tab:learnable}. The additional random anchors can bring about improvement. But as the improvement of $0.1\%$ NDS and $0.4\%$ mAP is small, we prefer not to use additional random anchors in our default setting.

\textbf{Effect of visual prompts for feature sharing.} To investigate the effects of adding a small number of prompt parameters to the feature maps of the backbone for feature sharing, we train a separate 2D detector that has its own image feature extraction backbone (denoted by ``No sharing"). We also train another 2D detector that directly uses the image features of the backbone trained by the 3D detector, and only the 2D detection head is fine-tuned (denoted by ``Sharing w/o prompt"). From the results in Table~\ref{tab:prompts}, we can see that ``No sharing" and ``Sharing w/o prompt" serve as the upper bound and lower bound of performance, respectively. 

The parameter $\tau$ in ``Stage 2" of Section ~\ref{subsec:optimize} controls the number of prompt parameters. We vary $\tau$ from 0.1 to 0.5 to study how the choice of it affects the performance. The results in Table~\ref{tab:prompts} show that all choices of $\tau$ can close the gap between ``No sharing" and ``Sharing w/o prompt", and $\tau=0.2$ is the best value. With $\tau=0.2$, the NDS difference between ``No sharing" and ``Sharing w/o prompt" is reduced from $0.6\%$ to $0.1\%$, and the mAP difference is reduced from $1.0\%$ to $0.5\%$.

\subsection{Visualization Results}
 Visual comparison between StreamPETR-8DQuery and StreamPETR-QAF2D are shown in Fig.~\ref{fig:visualization}. We draw 3D detection boxes in multi-camera images and their projections in BEV space. Three typical cases where the proposed QAF2D helps are given. Case 1 shows that QAF2D can remove the false positive (blue box in the purple ellipse in the top-left image of Fig.~\ref{fig:visualization} (a)) and make the true positive more accurate (see the alignment between the prediction and ground truth in BEV). Case 2 demonstrates that when the objects are faraway, the prediction of QAF2D is more accurate as well. Case 3 shows that when the object is small and difficult to distinguish from the background, QAF2D can help to alleviate miss detection. Please refer to the appendix for more visualization results.

\vspace{-2mm}
\section{Conclusion and Limitation}
In this paper, we propose to generate 3D query anchors from 2D boxes so that the more reliable 2D detection results can be used to boost the performance of 3D detectors. To share the image feature backbone between 2D and 3D detectors while keeping the performance of the 3D detector uncompromised, we design a two-stage optimization approach with visual prompts. We integrate the proposed approach into three query-based 3D object detectors, and comprehensive experiments are carried out on the nuScenes dataset to verify its effectiveness. 

A limitation of our approach is that 3D detection results depend on the quality of 2D detectors (though not sensitive to it). If the 2D detector misses an object, it should be difficult for a query-based 3D detector to recover the missed object. Meanwhile, combining the 3D anchors generated by our approach with the random anchors in a straightforward manner does not produce notable improvement. We will investigate how to achieve synergy between the two kinds of anchors in our future work.   

\noindent{\textbf{Acknowledgement.} P. Liang was supported in part by a Fundamental Research Cultivation Fund of ZZU.}

{
    \small
    \bibliographystyle{ieeenat_fullname}
    \bibliography{main}
}

\clearpage

\section*{Appendix}

\section{More Implementation Details}
\label{sec:moredetails}
We present the width range ($w_g^\mathrm{min}$, $w_g^\mathrm{max}$), height range ($h_g^\mathrm{min}$, $h_g^\mathrm{max}$), and length range ($l_g^\mathrm{min}$, $l_g^\mathrm{max}$) of each category $g$ in Table~\ref{tab:range}. The interval used to sample width, height, and length candidates is 0.05m.


\begin{table}[h]
\centering
\resizebox{1.0\linewidth}{!}{
\setlength{\tabcolsep}{.1em}%
\begin{tabular}{c|c|c|c}
\hline
Category & ($w_g^\mathrm{min}$, $w_g^\mathrm{max}$) & ($h_g^\mathrm{min}$, $h_g^\mathrm{max}$) & ($l_g^\mathrm{min}$, $l_g^\mathrm{max}$) \\ \hline
Car                       & (1.4, 2.8)         & (1.2, 3.1)          & (3.4, 6.6)         \\ \hline
Pedestrian                & (0.3, 1.0)         & (1.0, 2.2)          & (0.3, 1.3)         \\ \hline
Bus                       & (2.6, 3.5)         & (2.8, 4.6)          & (6.9, 13.8)        \\ \hline
Truck                     & (1.7, 3.5)         & (1.7, 4.5)          & (4.5, 14.0)        \\ \hline
Trailer                   & (2.2, 2.3)         & (3.3, 3.9)          & (1.7, 14.0)        \\ \hline
\begin{tabular}[c]{@{}c@{}}Construction \\ vehicle\end{tabular}     & (2.1, 3.4)         & (2.0, 3.0)          & (3.7, 7.6)        \\ \hline
Motorcycle                 & (0.4, 1.5)        & (1.1, 2.0)         & (1.2, 2.8)         \\ \hline
Bicycle                   & (0.4, 0.9)         & (0.9, 2.0)         & (1.3, 2.0)         \\ \hline
Traffic cone             & (0.2, 1.2)        & (0.5, 1.4)          & (1.3, 2.0)         \\ \hline
Barrier                   & (1.7, 3.6)         & (0.8, 1.4)          & (0.3, 0.8)         \\ \hline
\end{tabular}
}
\caption{The width, height, and length ranges of each class for anchor generation. The unit is meter.}
 \vspace{-2mm}
\label{tab:range}
\end{table}

\begin{figure*}[h]
  \centering
   \includegraphics[width=0.95\linewidth]{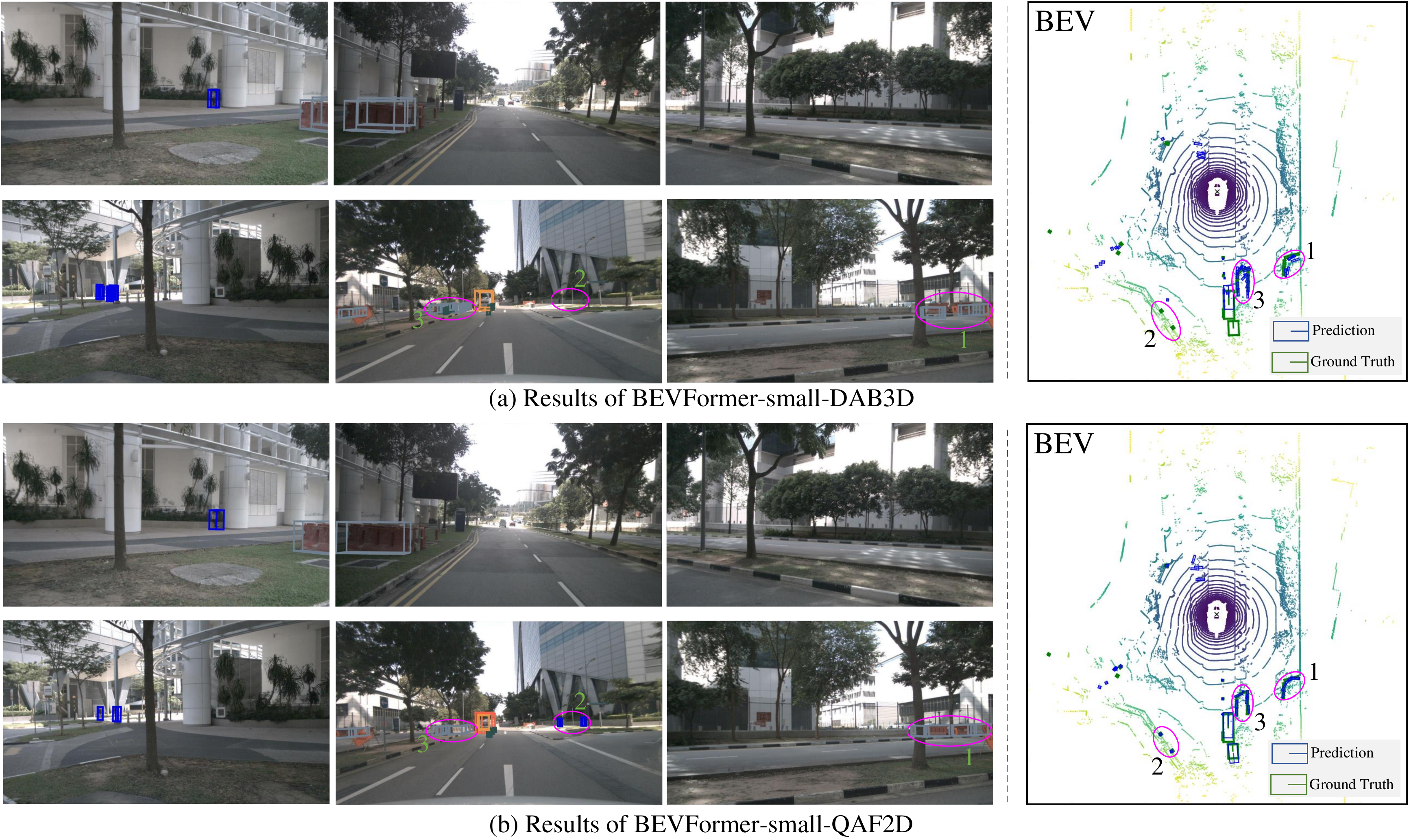}
   \vspace{-2.mm}
   \caption{Visualization results of BEVFormer-small-DAB3D and  BEVFormer-small-QAF2D. The results in multi-camera images are shown on the
left, and the corresponding results in bird’s-eye-view are shown on the right.}
   \label{fig:bevformer}
   \vspace{-3mm}
\end{figure*}



\section{More Visualization Results}
Fig.~\ref{fig:bevformer} shows the visual comparison between BEVFormer-small-DAB3D~\cite{eccv2022bev} and its QAF2D-enhanced version.  Case 1 and Case 3 show that QAF2D can make the detections more accurate. Case 2 demonstrates that QAF2D can help detect small objects that are missed by BEVFormer-small-DAB3D.

\begin{figure*}[h]
  \centering
   \includegraphics[width=0.95\linewidth]{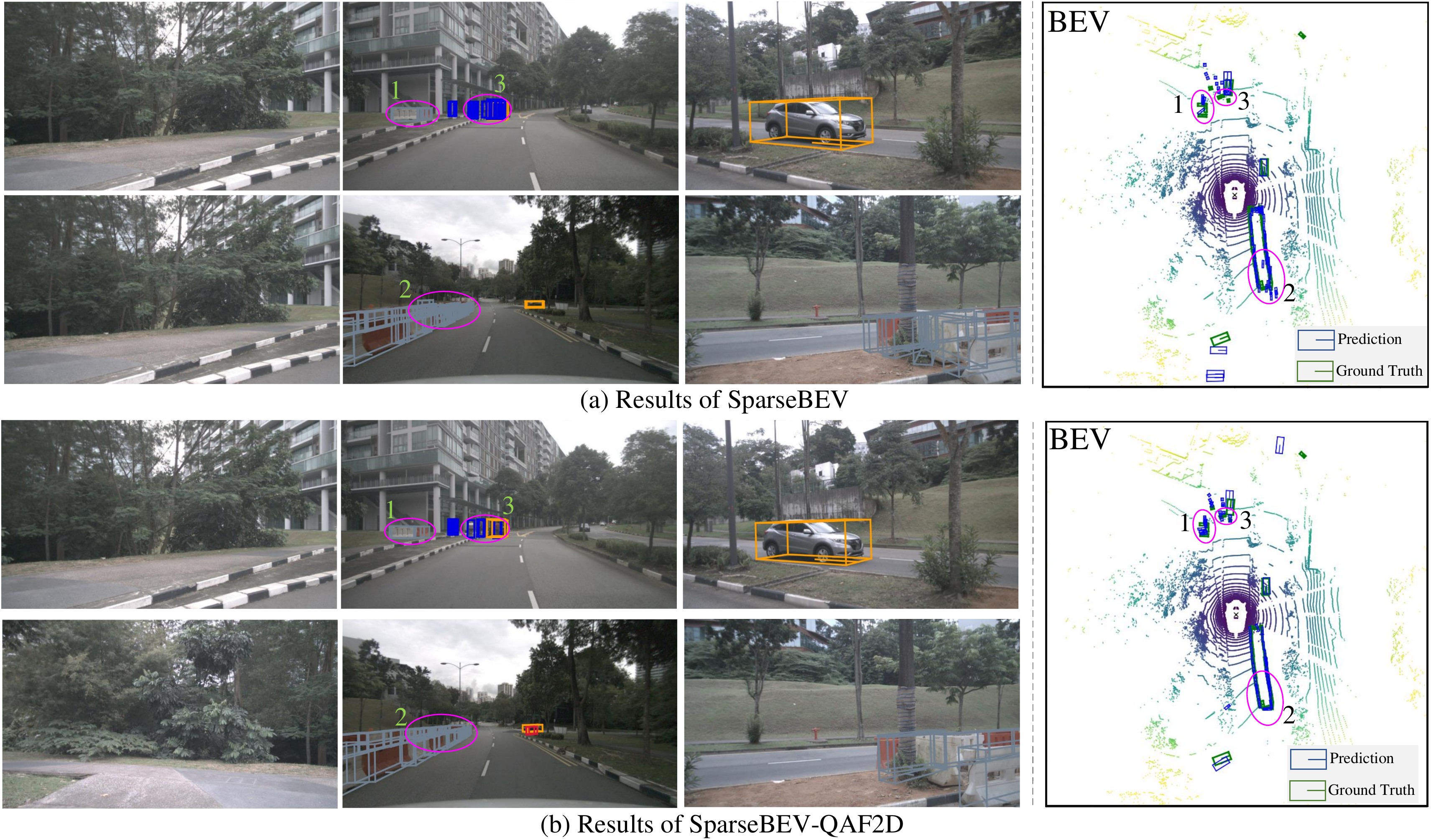}
   \vspace{-2.mm}
   \caption{Visualization results of SparseBEV and  SparseBEV-QAF2D. The results in multi-camera images are shown on the
left, and the corresponding results in bird’s-eye-view are shown on the right.}
   \label{fig:sparsebev}
   \vspace{-3mm}
\end{figure*}

Fig.~\ref{fig:sparsebev} shows the visual comparison between SparseBEV~\cite{iccv2023sparsebev} and its QAF2D-enhanced version. Case 1 and Case 2 show that QAF2D is useful in improving the accuracy of detection results, and Case 3 demonstrates that QAF2D can alleviate the problem of missed detection of small objects.





\end{document}